\documentclass[10pt,twocolumn,letterpaper]{article}

\usepackage{cvpr}
\usepackage{times}
\usepackage{graphicx}

\usepackage{amsmath}
\usepackage{amssymb}
\usepackage[super]{nth}
\usepackage{float}
\usepackage{upgreek}
\usepackage{newtxtext,newtxmath}

\usepackage[pagebackref=true,breaklinks=true,letterpaper=true,colorlinks,bookmarks=false]{hyperref}
\cvprfinalcopy 
\ifcvprfinal\fi

\begin{document}

\title{Event-based High Dynamic Range Image and Very High Frame Rate Video Generation using Conditional Generative Adversarial Networks}
\author{S. Mohammad Mostafavi I.$^{1}$*, Lin Wang $^{2}$\thanks{These two authors contributed equally}
, Yo-Sung Ho$^{1}$, Kuk-Jin Yoon$^{2}$\\
$^{1}$Gwangju Institute of Science and Technology (GIST), \\
$^{2}$Korea Advanced Institute of Science and Technology (KAIST)\\
{\tt\small mostafavi@gist.ac.kr, wanglin@kaist.ac.kr, hoyo@gist.ac.kr, kjyoon@kaist.ac.kr}
}
\maketitle

\begin{abstract}
    
    Event cameras have a lot of advantages over traditional cameras, such as low latency, high temporal resolution, and high dynamic range. 
    However, since the outputs of event cameras are the sequences of asynchronous events over time rather than actual intensity images, existing algorithms could not be directly applied. Therefore, it is demanding to generate intensity images from events for other tasks. 
    %
   %
    In this paper, we unlock the potential of event camera-based conditional generative adversarial networks to create images/videos from an adjustable portion of the event data stream. 
    The stacks of space-time coordinates of events are used as inputs and the network is trained to reproduce images based on the spatio-temporal intensity changes. 
    The usefulness of event cameras to generate high dynamic range (HDR) images even in extreme illumination conditions and also non blurred images under rapid motion is also shown. 
    In addition, the possibility of generating very high frame rate videos is demonstrated, theoretically up to 1 million frames per second (FPS) since the temporal resolution of event cameras are about 1 $\mu$s. 
    %
    Proposed methods are evaluated by comparing the results with the intensity images captured on the same pixel grid-line of  events using online available real datasets and synthetic datasets produced by the event camera simulator. 
  
\end{abstract}
\section{Introduction}
Event cameras are bio-inspired vision sensors that mimic the human eye in receiving the visual information \cite{lichtsteiner2008128}. 
While traditional cameras transmit intensity frames at a fixed rate, event cameras transmit the changes of intensity 
at the time of the changes, in the form of asynchronous events that deliver space-time coordinates of the intensity changes. 
They have lots of advantages over traditional cameras, \eg low latency in the order of microseconds, high temporal resolution (around 1 $\upmu$s) and high dynamic range. 
However, since the outputs of events cameras are the sequences of asynchronous events over time rather than actual intensity images, most existing algorithms cannot be directly applied. Thus, although it has been recently shown that event cameras are sufficient to perform some tasks such as 6-DoF pose estimation\cite{rebecq2017evo} and 3D reconstruction \cite{nguyenreal,kim2016real}, it will be a great help if we can generate intensity images from events for other tasks such as object detection, tracking and SLAM. 

\newcommand{\bsize}{0.145} 
\begin{figure}[t]
\begin{center}
\includegraphics[width=\bsize\textwidth]{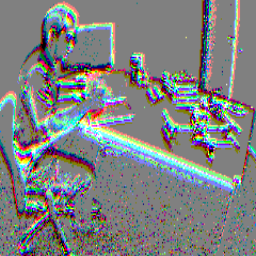}
\includegraphics[width=\bsize\textwidth]{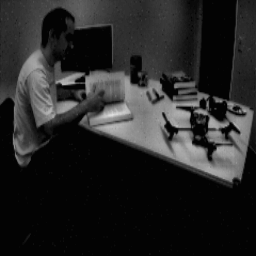}
\includegraphics[width=\bsize\textwidth]{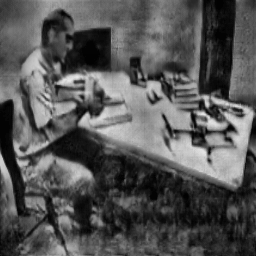}
\includegraphics[width=\bsize\textwidth]{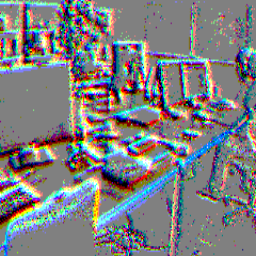}
\includegraphics[width=\bsize\textwidth]{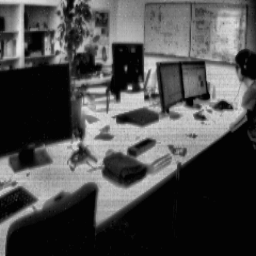}
\includegraphics[width=\bsize\textwidth]{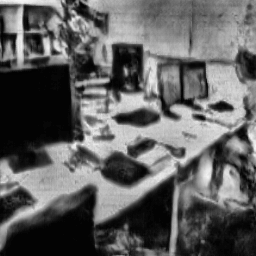}
\end{center}
\vspace{-8pt}
   \caption{From left to right, input events, active pixel sensor (APS) images from the DAVIS camera, and our results. 
   Our methods construct HDR images with more details that normal cameras could not reproduce as in APS frames. 
   We will show high frame rate video generation results in the supplementary material.}
   \vspace{-6pt}
\label{fig:impress}
\end{figure}

Actually, it has been stated that event cameras, in principle, transfer all the information needed to reconstruct images or a full video stream \cite{bardow2016simultaneous,reinbacher2016real,rebecq2017evo}. However, this statement has never been thoroughly substantiated. 
Motivated by recent advances of deep learning in image reconstruction and translation, we tackle the problem of generating intensity images from events, 
and further unlock the potential of event cameras to produce high quality HDR intensity images and high frame rate videos with no motion blur, 
which is especially important when the  robustness to fast motion and to extreme illumination conditions is critical as in autonomous driving. 
To the best of our knowledge, our work is the first attempt focusing on pure events to HDR images and high frame rate video translation, and proving that event cameras can produce high-quality non-blurred images and videos even under fast motion and extreme illumination conditions.   
We first propose the event-based domain translation framework that generates better quality images from events compared with active pixel sensor (APS) frames and other previous methods. For this framework, two novel and initiative event stacking methods are also proposed based on shifting over the event stream, stacking based on time (SBT) and stacking based on the number of events (SBE), such that we can reach high frame rate and HDR representation with no motion blur, which is, in contrast, impossible for the normal cameras. It turns out that it is possible to generate a video with up to 1 million FPS using these stacking methods. 

To verify the robustness of the proposed methods, we conduct intensive experiments and evaluation/comparison. In experiments, real datasets from a dynamic and active-pixel vision sensor, DAVIS, which is a joint event and intensity camera \cite{mueggler2017event}, are used. The sensor's pixel grid-line of the events and the intensity are on the same location which helps reducing extra steps of rectification and warping for adjusting two images to each other. We make an open dataset that includes more than 17$K$ images captured by the DAVIS camera to learn a generic model for event-to-image/video translation. In addition, we make a synthetic dataset containing 17$K$ images by using the event camera simulator \cite{rebecq2018esim} for experiments. 

\section{Related work}

\subsection{Intensity-image reconstruction from events}
One of the early attempts on visually interpreting or reconstructing the intensity image from events is the work by Cook \etal  \cite{cook2011interacting}, in which recurrently interconnected areas called maps were utilized to interpret intensity and optic flow. The model guides the network of relations between the maps of optical flow, intensity, spatial and temporal intensity derivative, camera calibration, and the 3D rotation to converge towards a global mutual consistency. Kim \etal  \cite{kim2008simultaneous} used pure events on rotation only scenes to track the camera and also built a super-resolution accurate mosaic of the scene based on probabilistic filtering. In \cite{barua2016direct}, intensity images were  reconstructed using a patch-based sparse dictionary both on simulated and real event data in the presence of noise. 
Bardow \etal \cite{bardow2016simultaneous} took a few steps further by reconstructing the intensity image and the motion field for generic motion in contrast to previous rotation only schemes. They proposed to minimize a cost function defined with events and spatiotemporal regularization terms on a sliding window interval of the event stream. Moreover, they reached a near real-time implementation on GPU. 
Meanwhile, Reinbacher \etal \cite{reinbacher2016real} introduced a variational denoising framework that iteratively filters  incoming events. They guided the events through a manifold regarding their timestamps to reconstruct the image. In comparison to \cite{bardow2016simultaneous}, their method yields more grayscale variations in untextured areas and recovers more details, and their GPU based algorithm can also perform in real-time. The measurements and simulations on the event camera with RGBW color filters were proposed by Moeys \etal in \cite{moeys2017color}. They presented the naive and computational methods for reconstructing the intensity image. The former requires an initial APS image from the event camera and updates the image with the incoming events, but does not produce sharp edges and background noise has negative effect on the outputs. The latter, on the other hand, creates better results based on an iterative scheme that creates a regularized image by solving the Poisson equation about the divergence of the intensity image and can run in real-time.

The aforementioned methods did create intensity images mainly by pure events, however, the reconstruction was not photorealistic. Recently, Shedligeri \etal \cite{shedligeri2018photorealistic} introduced a hybrid method that fuses intensity images and events to create photorealistic images. Their method relies on a set of three autoencoders.  
This method produces promising results for normally illuminated scenes, but it fails in recovering HDR scenes under extreme illumination conditions since it only utilizes event data for finding the 6-DoF pose.

\subsection{Deep learning on events}
Although deep learning has not been much applied to event-based vision, some recent studies have demonstrated that deep learning successfully performs with event data. 
Moeys \etal~\cite{moeys2016steering} utilized both event data and APS images to train a convolutional neural network (CNN) for controlling the steering of a predator robot. Other methods on steering prediction for self-driving cars by using pure events and/or by incorporating the APS images in an end-to-end fashion have been also studied in \cite{binas2017ddd17,maqueda2018event}. 
On the other hand, a stacked spatial LSTM network was introduced in \cite{nguyenreal}, which relocalizes the 6-DoF pose from events, and the  optical flow estimation based on a self-supervised encoder-decoder network was proposed in \cite{zhu2018ev}. 

Supervised learning is adopted to create pseudo labels for detecting objects under ego-motion in \cite{chen2018pseudo}. The pseudo labels are transferred to the event image by training a CNN on APS images. And, as mentioned in the previous section, the fusion of event data and APS images was introduced in  \cite{shedligeri2018photorealistic}, which utilized autoencoders to create photorealistic images. To the best of our knowledge, we are the first to apply generative adversarial networks on event data.

\subsection{Condition GANs on image translation}
Actually, there is no qualitative research showing the effectiveness of conditional GANs (cGANs) on event data. Prior works have focused on cGANs for image prediction from a normal map\cite{wang2016generative}, future frame prediction\cite{mathieu2015deep} and image generation from sparse annotations\cite{karacan2016learning}. The difference between using GANs for image-to-image translation conditionally and unconditionally is that unconditional GANs highly rely on the confining lost function to control the output to be conditioned. cGANs have been successfully applied to style transfer \cite{li2016precomputed, atapour2018real,isola2017image, zhu2017unpaired,ledig2017photo} in the frame image domain, and these applications mostly focused on converting an image from one representation to another based on the supervised setting. Besides, it requires input-output pairs for graphics tasks while assuming some relationship between domains. When comes to event vision, cGANs have not yet been examined qualitatively and quantitatively, and therefore, we seek to unlock the potential of cGANs for image reconstruction based on event data. However, since the general approach for frame-based image translation is typically different from event-based one, we first propose a deep learning framework to accomplish this task and fully take advantages of an event camera such as low latency, high temporal resolution, high dynamic range with the proposed framework. We then qualitatively and quantitatively evaluate the proposed framework with real and synthetic datasets. 

\section{Proposed method}  
To reconstruct HDR and high temporal resolution images and videos from events, we exploit currently available deep learning models, such as cGANs, as potential solutions for event vision.
cGANs are generative models that learn a mapping from observed image \emph{x} and random noise vector \emph{z} to the output image $y$, $G:\{x, z\}{\rightarrow}y$. The generator $G$ is trained to produce output that is not distinguishable from original images by an adversarially trained discriminator, \emph{D} 
\cite{goodfellow2014generative}.  
The objective is to minimize the distance between ground truth and output from generator, and to maximize the observation from discriminator.     

cGANs such as Pix2Pix \cite{isola2017image} and CycleGANs \cite{zhu2017unpaired} have proved their capability in image-to-image translation bringing breakthrough results. 
The key strength of cGANs is that there is no need to tailor the loss function regrading given specific tasks, 
and it can generally adapt its own learned loss to the data domain where it is trained. However, event data is quite different from those used for traditional vision approaches based on cGANs, so we propose new methods that can provide off-the-shelf inputs for neural networks in Sec. \ref{EventStacking} first and build a network in Sec. \ref{proposed_network}.

\subsection{Event stacking} \label{EventStacking}
\begin{figure*}[t]
\begin{center}
\vspace{0pt}
    \includegraphics[width=0.93\linewidth]{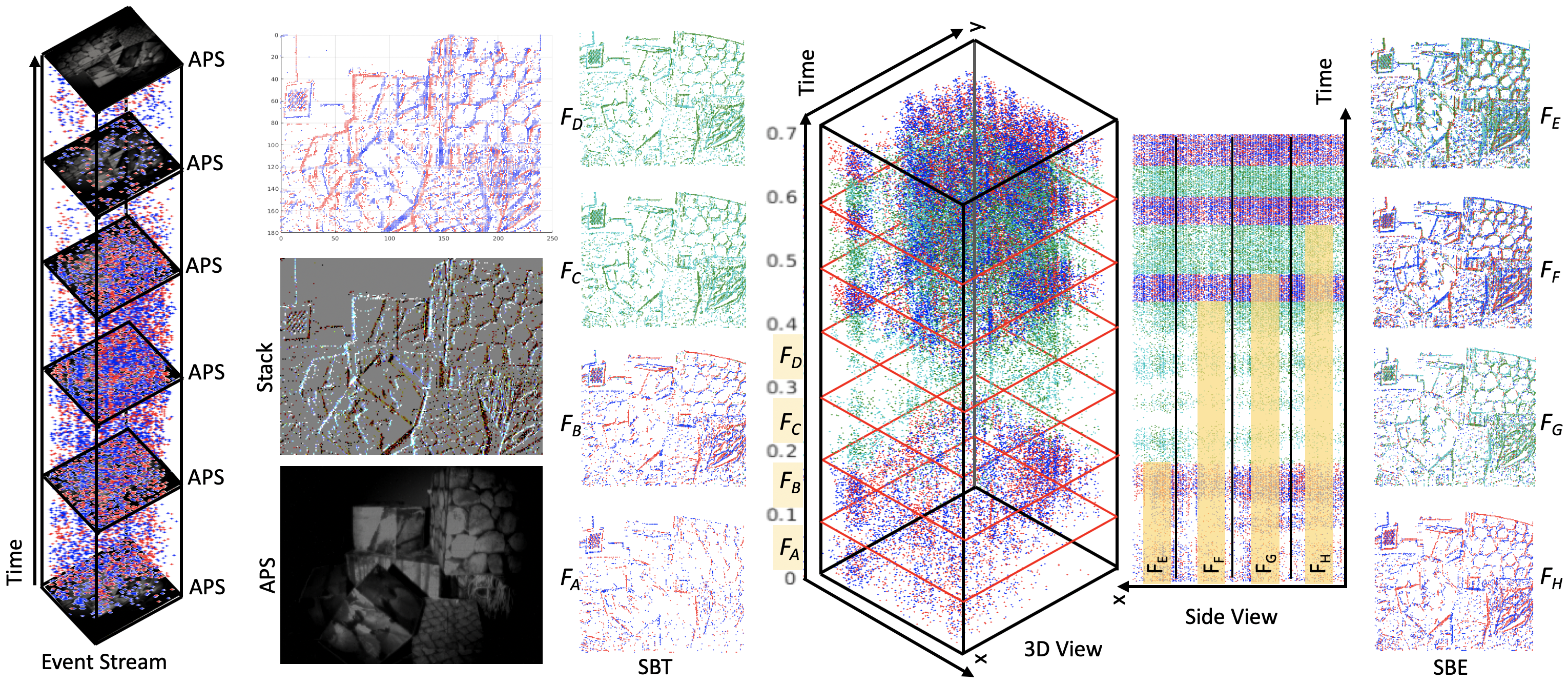}
    \caption{The event stream and construction of stacks by SBT and SBE. 
    Two main color tuples of (Red(+), Blue(-)) and (Green(+), Cyan(-)) express the event polarity (plus, minus) throughout this paper.
    In the main 3D view two types of stacking (SBT on left and SBE on right) are shown using the yellow highlighted time.
    The 3D view followed by its side view are color coded with (Red, Blue) and (Green, Cyan) periodically (every 5000 events) for better visualization.
    All the images and plotted data are from the "hdr\underline{ }boxes" sequence of \cite{mueggler2017event}.}
    \label{fig:stream}
\end{center}
\vspace{-15pt}
\end{figure*}

In an event camera, each event $e$ is represented as a tuple $(u, v, t, p)$, where $u$ and $v$ are the pixel coordinates and $t$ is the timestamp of the event, and $p=\pm 1$ is the polarity of the event, which is the sign of the brightness change ($p=0$ for no event). 
These events are shown as a stream on the left of  Fig.~\ref{fig:stream}. Based on the frame rate of intensity camera, we have synchronized APS images and asynchronous events in-between two consecutive APS frames. 
To feed event data input to the network, new representations of event data are required. One simple way is to form the 3D event volume 
as $p(u, v, t)$ for some time duration ensuring event data enough for image reconstruction. 
When denoting the temporal resolution of an event camera by $\delta t$ and the time duration by $t_d$, the size of the 3D volume is $(w, h, n)$, where $w$ and $h$ represent the spatial resolution of an event camera and $n=t_d/\delta t$. This is equivalent to have the $n$-channel image input for the network. This representation preserves all the information about events. However, the problem is that the number of channels is very huge. For example, when $t_d$ is set to 10$m$s, then $n$ is about 10$K$, which is extraordinarily large, since the temporal resolution of an event camera is about 1 $\mu$s.
For this reason, we construct the 3D event volume with small $n$ 
by forming each channel via merging and stacking the events within a small time interval. 
Event stacking can be done in different ways, but the temporal information of event is necessarily sacrificed in return. 


\vspace{-9pt}
\subsubsection{Stacking Based on Time (SBT)} 
\vspace{-4pt}
In this approach, the streaming events in-between the time references of two consecutive intensity images (APS) of the event camera, denoted as $\Delta t$, are merged. But not all events are merged into a single frame. Instead, the time duration of the event stream is devided into $n$ equal-scale portions, and then $n$ grayscale frames, $S^i_p(u, v)$, $i=1, 2,.., n$, are formed by merging the events in each time interval $[\frac{(i-1)\Delta t}{n}, \frac{i \Delta t}{n}]$. $S^i_p(u, v)$ is the sum of polarity($p$) values at $(u,v)$. These $n$ grayscale frames are stacked together again to form one stack $S_p(u, v, i)=S^i_p(u, v)$, $i=1, 2,.., n$, which is fed to the network as the input. As mentioned, this stacking method loses the time information of events within time interval $\frac{\Delta t}{n}$. However, the stack itself, as the sequence of frames from one to $n$, still holds the temporal information to some extent. 
Therefore, larger $n$ can keep more temporal information. 
%
%

Fig~\ref{fig:stream} illustrates how to merge and stack the events. When $n=3$ (\ie stacking frames $F_A$, $F_B$, and $F_C$ into one stack), the stack can be visualized as a pseudo color frame, as shown in the left part of Fig. \ref{fig:stream} above the APS image. 
Based on the time shown at the event manifold in the middle of Fig.~\ref{fig:stream}, starting from time zero on the 3D view, the location of APS image is around the location of the third red rectangle near 0.03 sec (the frame rate of the APS image is 33 FPS). 
\vspace{-9pt}
\subsubsection{Stacking Based on the number of Events (SBE)}
\vspace{-4pt}
Unfortunately, SBT brings an intrinsic limitation originated from the event camera, which is the lack of events when there is no movement of the scene or the camera. When the event data within the time interval are not enough for the image reconstruction, it is hard to get good HDR images inevitably. This is the case for the fourth and fifth  frame of the event stream at the left of Fig.~\ref{fig:stream}. Furthermore, another flaw comes from the case of having too many events in one time frame as in the third time frame.

SBE more coincides with the nature of an event camera, which is being asynchronous to time, and can overcome the aforementioned limitations of SBT. In this method, a frame is formed by merging the events based on the number of incoming events as illustrated in 
Fig.\ref{fig:stream}. 
The first $N_e$ events are merged into frame 1 and next $N_e$ events into frame 2, and this is continued up to frame $n$ to create one stack of $n$ frames. Then, this $n$-frame stack containing $n N_e$ events in total is used as an input to the network. This method guarantees rich event data enough to reconstruct images depending on the $N_e$ value. 
$F_E$, $F_F$, $F_G$, and $F_H$ in  Fig~\ref{fig:stream} are the frames corresponding to different numbers of events, $N_e, 2N_e, 3N_e, 4N_e$, respectively. 
%
%
Since we count the number of events with time, we can adaptively adjust the number of events in each frame and also in one stack. 

\vspace{-9pt}
\subsubsection{Stacking for video reconstruction}
\label{sec:video_stack}
\vspace{-4pt}

Both SBT and SBE can be applied for video reconstruction from events using the proposed network, and in both methods, the frame rate of the output video can be adjusted by controlling the amount of time shift of two adjacent event stacks used as inputs to the network.   
When the events in the time interval $[i-\Delta t, i]$ are used for one input stack for the image $I(i)$ in a video, the next input stack for the image $I(i+t_s)$ in a video can be constructed by using the events in the time interval $[i-\Delta t', i+t_s]$ (for SBT $\Delta t'=\Delta t-t_s$),  with the time shift $t_s$. Then, the frame rate of the output video becomes $\frac{1}{t_s}$.  
%
%
%
It is also worthy of notice that two stacks have large time overlap $[i-\Delta t', i]$ with   duration $\Delta t'$. If $\Delta t' >> t_s$, the temporal consistency is naturally enforced for nearby frames. Since the temporal resolution of an event camera is about 1 $\mu$s, we can reach up to one million FPS video with temporal consistency. This will be demonstrated in Sec.~\ref{sec:exp}



\subsection{Network architectures} \label{proposed_network} 
\begin{figure*}[t]
\centering
\includegraphics[width=\textwidth, scale=0.5]{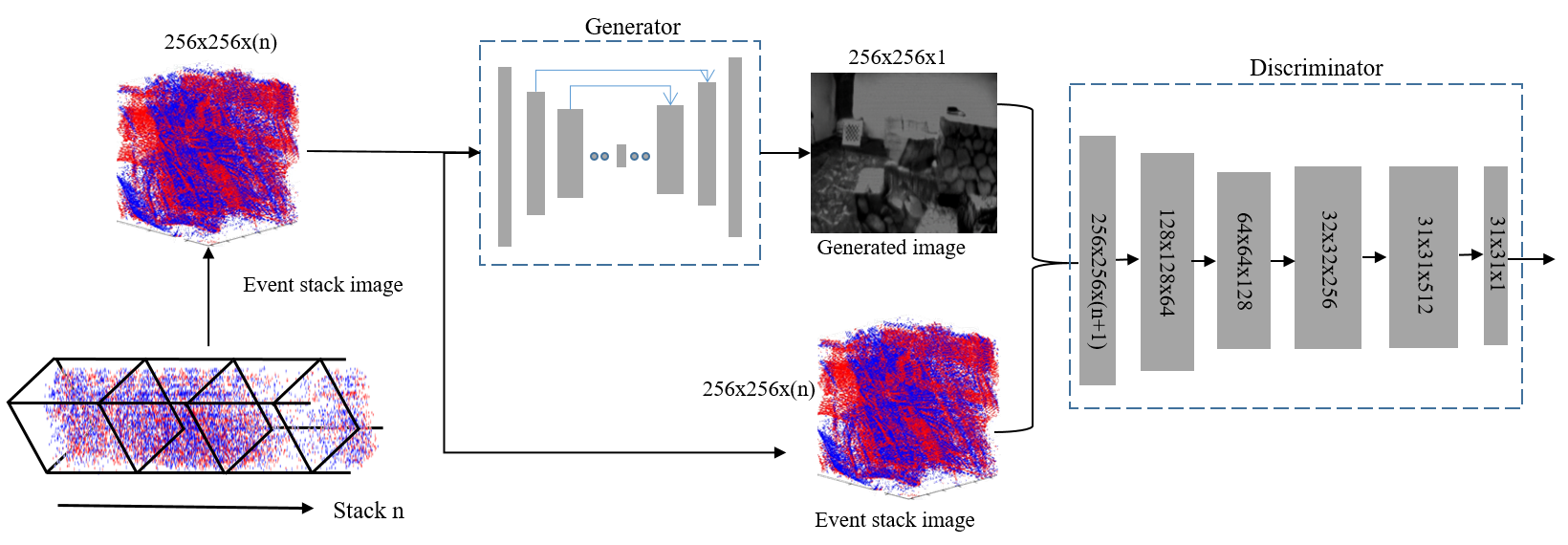}
\vspace{-13pt}
\caption{The proposed framework with the generator and discriminator networks. Our discriminator network is similar to PatchGAN \cite{yi2017dualgan}, which takes two images (original APS image and the image generated by the generator from events). The discriminator first concatenates the condition of feature maps from the last layer of the generator and discriminates whether the generated image respects the condition of domain transfer from event to intensity.}
\label{fig:Descriminator}
\vspace{-5pt}
\end{figure*}

In this paper, we describe our generator and discriminator motivated by \cite{li2016precomputed}. Details of the architectures including the size of each layer can be found in Fig.~\ref{fig:Gen} and Fig.~\ref{fig:Descriminator}.

\vspace{-9pt}
\subsubsection{Generator architecture}
\vspace{-4pt}

\begin{figure*}[t]
\centering
\includegraphics[width=0.9\textwidth]{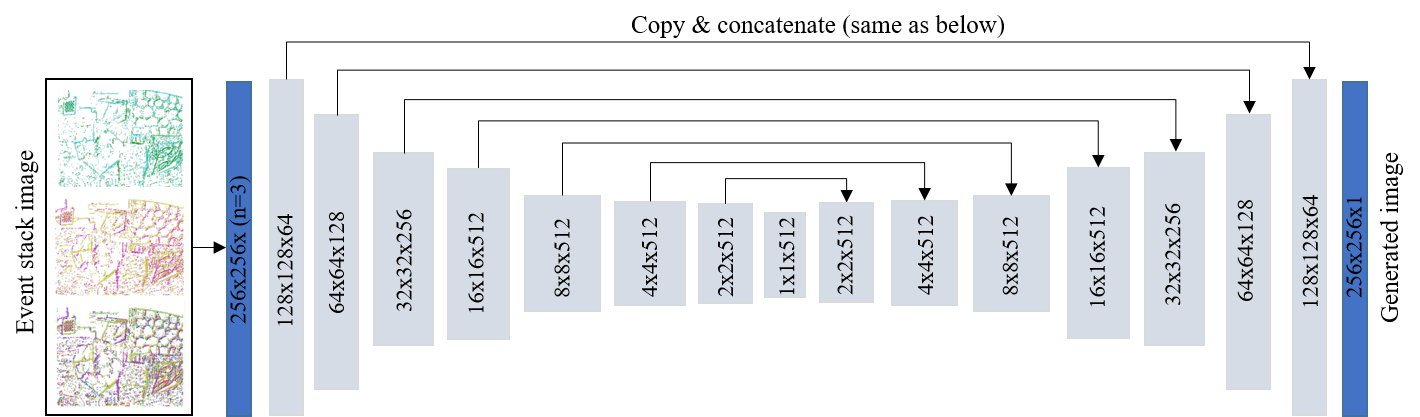}
\caption{Generator network: A U-network\cite{ronneberger2015u,isola2017image} architecture (with skip connections) that takes an input with the dimension of $256\times 256\times n$ ($n=3$ for this example), followed by gray boxes corresponding to multi-channel feature maps. The number of channels is denoted inside each box. The first two numbers(from bottom to top) indicate the filter sizes and the last number indicates the number of filters. 
}
\vspace{-5pt}
\label{fig:Gen}
\end{figure*}

The core of the event-to-image translation is how to map a sparse event input to a dense HDR output with details, 
sharing the same structural image features, such as edges, corners, blobs, etc. 
Encoder-Decoder network is the mostly used network for image to image translation tasks. The input is continuously downsampled through the network, and then upsampled back to get the translated result. Since, in the event-to-image translation problem, there is a huge amount of high-frequency important information from event data
passing through the network, it is likely to lose detailed features of events during this process and induce noise to the outputs. 
For that reason, we consider the similar approaches proposed in \cite{isola2017image}, where we further add skip connections to the "U-net" network structure in \cite{reinbacher2016real}. In Fig. \ref{fig:Gen}, the detailed information including number of layers and inputs/output are depicted.  

\vspace{-9pt}
\subsubsection{Discriminator architecture}
\vspace{-4pt}

The function of the discriminator is to classify a generated image from events as real or fake. In other words, the generator is trying to maximize the chance of the discriminator misclassifying the image from events and the discriminator is, in turn, trying to maximize its chances of correctly classifying the incoming generated image. 

Our network is originated from the network in \cite{yi2017dualgan}. Fig~\ref{fig:Descriminator} illustrates the details of our network architecture. 
Our discriminator can be considered as a method to minimize the style transfer loss between events and intensity images.
Mathematically, the objective function is defined as
\begin{equation}
\begin{split}
    L_{eGAN}(G,D)=E_{e,g}[\log{D(e,g)]} +\\
    E_{e,\epsilon}[\log{(1-D(G(e,\epsilon)))}].
\end{split}
\end{equation}
where $e$ indicates the original event, $g$ indicates the generated image, and $\epsilon$  indicates the Gaussian noise as input to the generator. Meanwhile, \emph{G} tries to minimize the difference of images from events, and \emph{D} is to maximize it. Here, for the regularization, the $L^1$ norm is used to shrink blurring as 
   \begin{equation}
    L_{L1}(G)=E_{e,g,\epsilon}[\|g-G(e,\epsilon)\|_1.
   \end{equation}
This $L^1$ norm is aimed to make the discriminator more focus on high-frequency structure of generated images from events. Eventually, the objective is to estimate the total loss from event-to-image translation as
  \begin{equation} \label{eq:2}
    G^* =\arg\min_{G}\max_{D} [L_{eGAN}(G,D)+\lambda L_{L1}(G)],
\end{equation} 
where $\lambda$ is a parameter to adjust the learning rate. With the noise $\epsilon$, the network could learn a mapping from event $e$ and $\epsilon$ to $g$, which could match the distribution based on events and help to produce more deterministic outputs.

\subsection{Dataset preparation} \label{DataPrepare}

Our training and test datasets are prepared based on three folds of methods. We create the first group of datasets by referring to \cite{mueggler2017event}, where many real-world scenes are included. We also make the second group of datasets by ourselves for various training and test purposes and also for opening to public afterwards. The datasets are captured using DAVIS camera, and have many series of scenarios. The third type of datasets is generated from ESIM\cite{rebecq2018esim}, an open-source event camera simulator. 
The real datasets contain many different indoor and outdoor scenes captured with various rotations and translations of the DAVIS camera . Our training data consist of pairs of stacked events as explained in Sec. ~\ref{EventStacking} together with the APS frames from both the real-world scenes and the ground truth (GT) frames generated in ESIM. 
Here, to use real data for training the network, we carefully prepare the training data to refrain the network from learning improper properties of the APS frames. Actually, APS frames suffer from motion blur under fast motion, and also have limited dynamic range resulting in the loss of details as shown in Fig.~\ref{fig:impress}. 
Therefore, directly using the real APS frames as ground truth is not a good way for training the network, since our goal is to produce HDR images with less blur by fully exploiting the advantages of event cameras. 


%
For that reason, the events relevant to the black and white regions of the training data are removed from the input to make the network learn to generate HDR images from events. 
%
In addition, the APS images are classified as blurred and non-blurred based on BRISQUE scores (that will be explained later) and manual inspection, and we refrain from using the blurred APS images in the training set. 
The simulated sequences are mainly generated from ESIM, where events are produced while a virtual camera moves in all directions to capture different scenes in given images. Since the events and APS images are generated from a controlled simulation environment, the APS frames are counted directly as the ground truth for image reconstruction. Therefore, the aforementioned training data refinement is not required for simulated datasets.

\section{Experiments and evaluation}
\label{sec:exp}

To explore the capability of our method, we conduct intensive experiments on the 
datasets depicted in Sec.~\ref{DataPrepare}, and also use another open-source dataset with three real sequences (Face, jumping, and ball) \cite{bardow2016simultaneous} for comparison. 
We create a training dataset about 60$K$ event stacks with corresponding APS image pairs based on their precise timestamps, 
and test our method on both scenes with normal illumination and also HDR scenes. From both the real and simulated datasets, we randomly chose 1,000 APS or ground truth images with corresponding event stacks, not used in the training step, for testing. 
Here, it is worthy of notice that, since real datasets do not include ground truth images for training and testing, we use their APS images as ground truth for training purposes. However, the APS image itself suffers from motion blur and low dynamic range. Thus, using APS images might not be the best way for training and also for evaluating the results. For that reason, we prepare the training APS images as described in Sec.~\ref{DataPrepare}, and assess the results using the structure similarity (SSIM) \cite{wang2004image}, feature similarity (FSIM) \cite{zhang2011fsim} computed by comparing the results with APS images, as well as by using the no-reference quality measure. In order to reach a holistic measure of quality, especially when evaluating the quality of reconstruction of real datasets without ground truth, the Blind/Referenceless Image Spatial Quality Evaluator (BRISQUE) \cite{mittal2012no}, which utilizes normalized luminance coefficients to quantify the \emph{naturalness} in images, is applied. 

On the other hand, to assess the similarity between ground truth and generated images for synthetic datasets created using ESIM \cite{rebecq2018esim}, each ground truth is matched with the corresponding reconstructed image with the closest timestamp, as mentioned in \cite{scheerlinck2018continuous}. 
The SSIM, FSIM, and the peak signal-to-noise ratio (PSNR) are adopted to evaluate non-HDR scenes and scenes that we have reliable ground truth. 

\subsection{SBT versus SBE} \label{sec:SBT_SBE}

We compare two event stacking methods, SBT and SBE, using our real datasets. 17$K$ event stack-APS image pairs are used for training, where we set $\Delta t$ for SBT to 0.03s and the number of events in one stack to 60$K$ for SBE. To clearly see the effect of a stacking method, the number of frames ($n$) in one stack is set to 3 for both methods. 
\begin{figure}[t]
\begin{center}
\renewcommand{\tabcolsep}{1pt}
\begin{tabular}{@{}ccc@{}}
   \includegraphics[width=28mm,height=24mm]{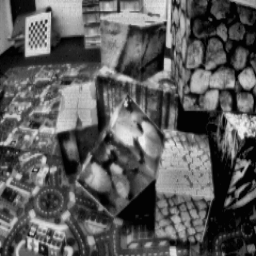} &
    \includegraphics[width=28mm,height=24mm]{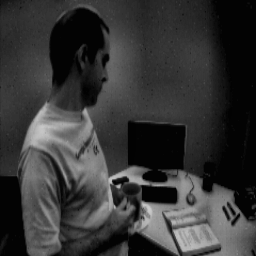} &
    \includegraphics[width=28mm,height=24mm]{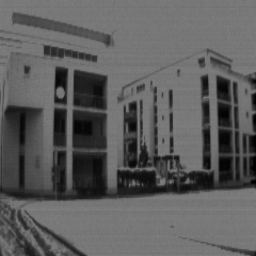} \\
    \includegraphics[width=28mm,height=24mm]{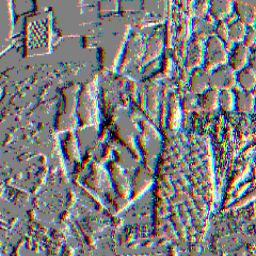} &
    \includegraphics[width=28mm,height=24mm]{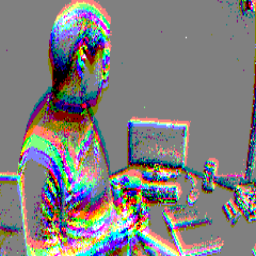} &
    \includegraphics[width=28mm,height=24mm]{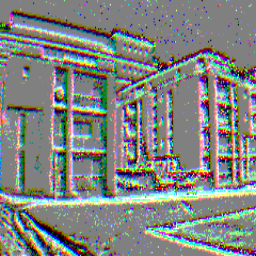} \\
    \includegraphics[width=28mm,height=24mm]{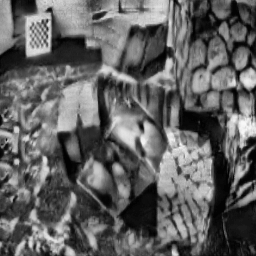} &
    \includegraphics[width=28mm,height=24mm]{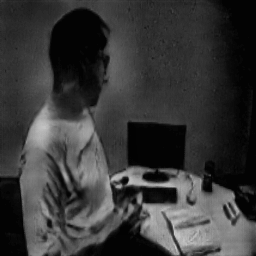} &
    \includegraphics[width=28mm,height=24mm]{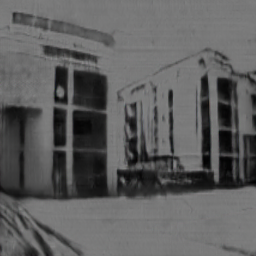} \\
    \includegraphics[width=28mm,height=24mm]{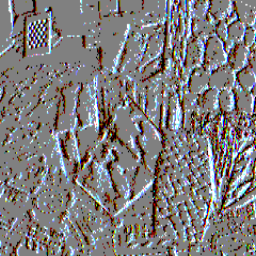} &
    \includegraphics[width=28mm,height=24mm]{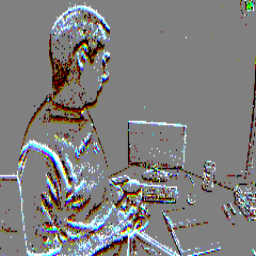} &
    \includegraphics[width=28mm,height=24mm]{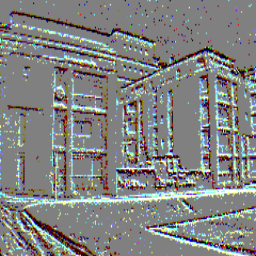} \\
    \includegraphics[width=28mm,height=24mm]{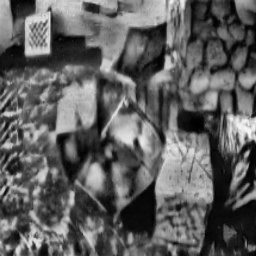} &
    \includegraphics[width=28mm,height=24mm]{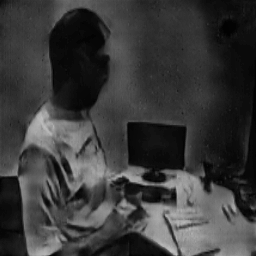} &
    \includegraphics[width=28mm,height=24mm]{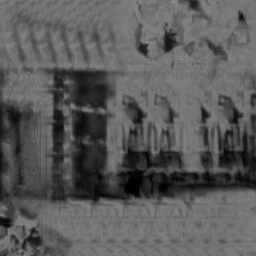} \\
\end{tabular}
\caption{Reconstruction results using input event stacks (visualized as pseudo color images) on different real-world sequences \cite{mueggler2017event}. From top to bottom,  APS images as ground truth, event stacks using SBE, reconstructed images with SBE, event stacks using SBT, and reconstructed images with SBT.}
\label{fig:compare}
\end{center}
\vspace{-18pt}
\end{figure}

Fig~\ref{fig:compare} shows reconstructed images on our real-world datasets using SBE and SBT, respectively, for qualitative comparison. It is shown that our methods (both SBT and SBE) are robust enough to reconstruct the images on different sequences, and the generated images are quite close to APS images considered as \emph{ground truth}. Our methods could successfully reconstruct shapes, appearance of human, building, etc. When comparing SBT and SBE, SBE produces better results in general. Table~\ref{table:realworldexp} shows quantitative evaluation results of using SBE. Note that large SSIM and FSIM values in  Table~\ref{table:realworldexp} do not always mean the better output quality because they just present the similarity with APS images suffering from motion blur and low dynamic range.  


\begin{table}
\caption{Quantitative evaluation of SBE on real-world datasets. 
}
\begin{center}
\vspace{-2pt}
\begin{tabular}{|l|c|c|c|}
\hline
 &BRISQUE & FSIM & SSIM\\
\hline\hline
Ours($n=3$) &37.79$\pm$5.86&0.85$\pm$0.05& 0.73$\pm$0.16 \\

\hline
\end{tabular}
\end{center}
\vspace{-10pt}
\label{table:realworldexp}
\end{table}

\subsection{Quantitative evaluation with simulated datasets}

In Sec.~\ref{sec:SBT_SBE}, we investigate the potential of our method on real-world data. Based on the results in Sec.~\ref{sec:SBT_SBE}, we find that SBE is more robust than SBT. Therefore, we conduct experiments based on SBE and show the robustness of our methods on datasets from ESIM \cite{rebecq2018esim}, which can generate large amount of reliable event data. Since the simulator produces noise-free APS images with corresponding events for a given image, APS images can be regarded as ground truth, leading to evaluate the results quantitatively. 
In addition, although our method is capable of stacking, namely, any number of frames ($n$) into a stack, we choose the number of channels $n=\{1,3\}$ to examine the effect of different numbers of channels. 
The number of events in one stack is set to 60$K$. 

\begin{figure*}[t]
\begin{center}
\renewcommand{\tabcolsep}{1pt}
\begin{tabular}{@{}cccccc@{}}
 &
    \includegraphics[width=28mm,height=24mm]{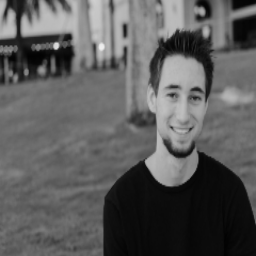} & 
    \includegraphics[width=28mm,height=24mm]{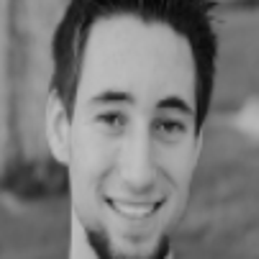} & &
    \includegraphics[width=28mm,height=24mm]{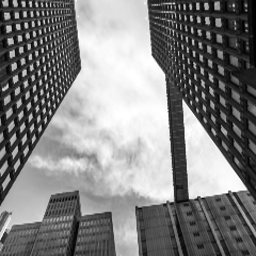} &
    \includegraphics[width=28mm,height=24mm]{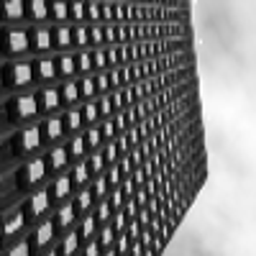} \\
    \includegraphics[width=28mm,height=24mm]{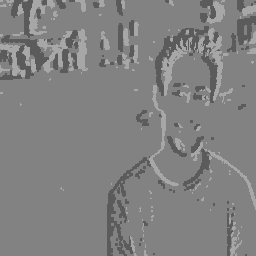} &
    \includegraphics[width=28mm,height=24mm]{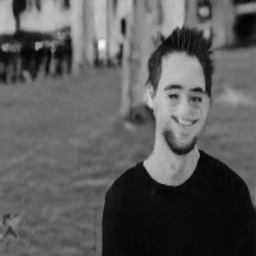} &
    \includegraphics[width=28mm,height=24mm]{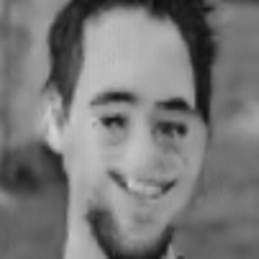} &
    \includegraphics[width=28mm,height=24mm]{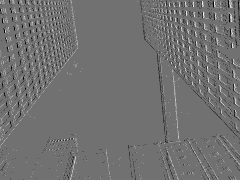} &
    \includegraphics[width=28mm,height=24mm]{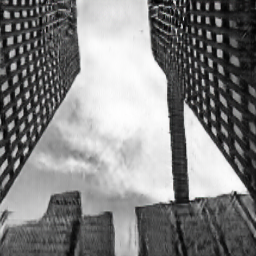} &
    \includegraphics[width=28mm,height=24mm]{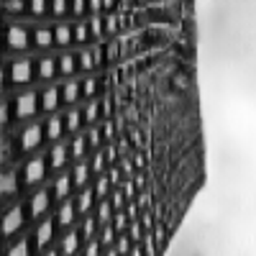} \\
   \includegraphics[width=28mm, height=24mm]{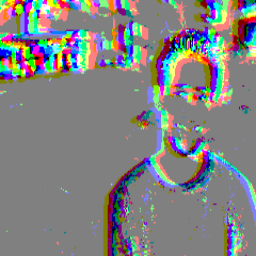} &
    \includegraphics[width=28mm, height=24mm]{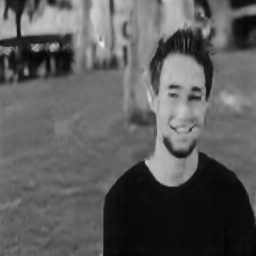} &
    \includegraphics[width=28mm,height=24mm]{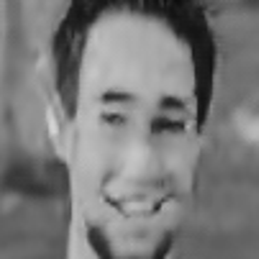} &
    \includegraphics[width=28mm,height=24mm]{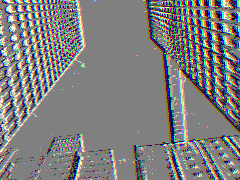} &
    \includegraphics[width=28mm, height=24mm]{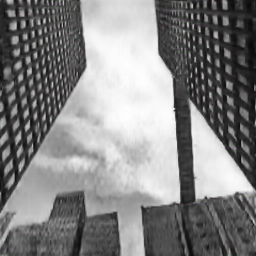} & 
    \includegraphics[width=28mm, height=24mm]{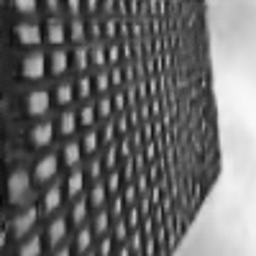} \\
\end{tabular}
\vspace{1pt}
\caption{Reconstructed outputs from the inputs generated by ESIM \cite{rebecq2018esim}. The first row shows the ground truth, and the second row shows the input events and reconstructed images using 1 frame per stack ($n=1$): images are distorted due to over-accumulated events. The third row shows the input events and reconstructed images using using 3 frames per stack ($n=3$), which is more robust than the one-frame stack.}
\label{fig:sim_Res}
\end{center}
\vspace{-10pt}
\end{figure*}

Table~\ref{table:simexp} shows the quantitative evaluation of our method with $n=1$ and $n=3$. 
It is shown that our method with $n=3$ produces better results than with $n=1$, proving that having more frames in one stack really improves the performance since it can preserve more temporal information as mentioned in Sec.~\ref{EventStacking}.  
In Fig.~\ref{fig:sim_Res}, we show a few reconstructed images as well as input event stacks and ground truth images. One thing needs to mention is that the face reconstructed with $n=1$ and the top of the building are a little bit distorted, which may be induced by too many events accumulated in one single channel. 

\begin{table}
\caption{Experiments on ESIM (simulator) datasets. 
Having more frames in one stack yields better results.}
\vspace{-5pt}
\begin{center}
\begin{tabular}{|l|c|c|c|c|}
\hline
 & PSNR (dB) & FSIM & SSIM \\
\hline\hline
Ours($n=1$) & 20.51$\pm$2.86 &  0.81$\pm$0.09  & 0.67$\pm$0.20 \\
Ours($n=3$)  & \textbf{24.87}$\pm$3.15 &  \textbf{0.87}$\pm$0.06  & \textbf{0.79}$\pm$0.12 \\
\hline
\end{tabular}
\end{center}
\vspace{-12pt}
\label{table:simexp}
\end{table}

\subsection{Comparison to relevant works}

\begin{figure*}[t]
\begin{center}
\renewcommand{\tabcolsep}{1pt}
\begin{tabular}{@{}cccccc@{}}
  \includegraphics[width=28mm,height=28mm]{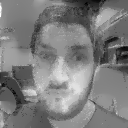} &
  \includegraphics[width=28mm,height=28mm]{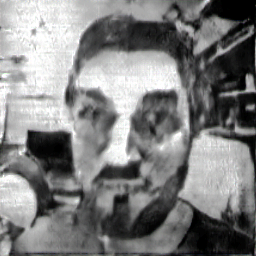}&
  \includegraphics[width=28mm,height=28mm]{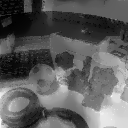} &
  \includegraphics[width=28mm,height=28mm]{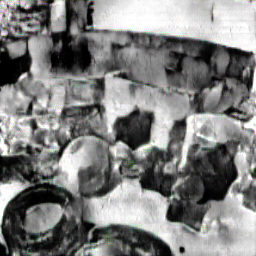} &
  \includegraphics[width=28mm,height=28mm]{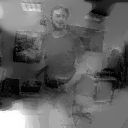} &
  \includegraphics[width=28mm,height=28mm]{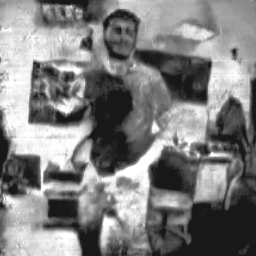} \\
\end{tabular}
\end{center}
\vspace{-8pt}
\caption{Comparison to the methods of Reinbacher \etal \cite{reinbacher2016real} and Munda \etal \cite{munda2018real}, which both utilize the dataset from Bardow \etal \cite{bardow2016simultaneous}. Since Reinbacher \etal did not open their source codes, we directly get the results from Munda \etal \cite{munda2018real}. The odd-number images are the results from Bardow \etal \cite{bardow2016simultaneous}, and the even-number images are the results of our method. We can easily see that our method produces more details ( e.g. face, beard, jumping pose, etc) as well as more natural gray variations in less textured areas.}
\vspace{-10pt}
\label{fig:compare_relatedworks}
\end{figure*}

We also qualitatively compare our methods on the sequences (\emph{face, jumping and ball}) with the results of manifold regularization (MR) \cite{munda2018real} and intensity estimation (IE) \cite{bardow2016simultaneous} in Fig~\ref{fig:compare_relatedworks}. Since we deal with highly dynamic data, 
we provide more persuasive and explicit explanation and results in the supplementary video, which shows the whole sequence of several hundred of frames. 

To compare the performance quantitatively, we use the BRISQUE score because no ground-truth image is available for these sequences. We compare the outputs of our method (SBE, $n=3$) on sequences (\emph{face, jumping and ball}) to the results of MR \cite{munda2018real} and IE \cite{bardow2016simultaneous} in Table~\ref{table:compareworks}. 
The results are quite consistent to the visual impression of Fig.~\ref{fig:compare_relatedworks}. Our outputs on all face, jumping, and ball sequences show much more details and result in relatively higher BRISQUE score.

\begin{table}
\caption{Quantitative comparison of our method to the methods in \cite{bardow2016simultaneous} and \cite{munda2018real}. The reported numbers are the mean and standard deviation of the BRISQUE measure applied to all reconstructed frames of the sequences. Our method shows better BRISQUE scores for all sequences.}
\vspace{-5pt}
\begin{center}
\begin{tabular}{|l|c|c|c|}
\hline
Sequence & Face & Jumping & Ball \\
\hline\hline
Bardow \cite{bardow2016simultaneous} & 22.27$\pm{8.81}$ & 29.39$\pm${7.27}  &29.37$\pm${9.61} \\
Munda \cite{munda2018real}  & 27.29$\pm${7.27} & 48.18$\pm${6.70} &34.98$\pm${9.31} \\
Ours($n=3$) & \textbf{48.26}$\pm${3.14} & \textbf{48.34}$\pm${2.18}  &  \textbf{39.18}$\pm{3.49}$ \\
\hline
\end{tabular}
\end{center}
\vspace{-15pt}
\label{table:compareworks}
\end{table}

\section{Discussion}

Although creating intensity images from an event stream itself is challenging, the resultant images 
can also be used for other vision tasks such as object recognition, tracking, 3D reconstruction, self-driving, SLAM etc. In that sense, the proposed method can be applied to many applications that use event cameras. Here, since the proposed method can fully exploit the advantages of events cameras such as high temporal resolution and high dynamic range, it can generate HDR images even better than APS images and very high frame rate videos as mentioned in Sec.~\ref {sec:video_stack}, greatly increasing the usefulness of the proposed method. 

%



\begin{figure}[tb]
\begin{center}
\renewcommand{\tabcolsep}{1pt}
\begin{tabular}{@{}ccc@{}}
\includegraphics[width=28mm,height=19mm]{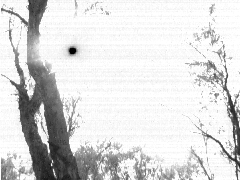}&
\includegraphics[width=28mm,height=19mm]{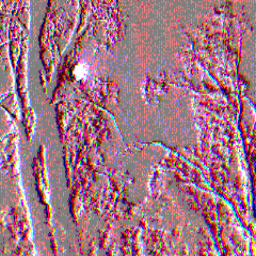} &
\includegraphics[width=28mm,height=19mm]{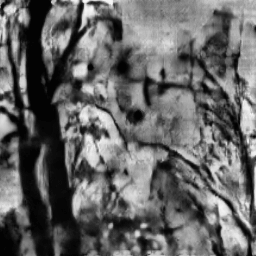}\\
\end{tabular}
\caption{HDR imaging against direct sunlight (extreme illumination). Left to right: APS, event stack, our reconstruction result. 
(sequence from \cite{scheerlinck2018continuous}).
}
\label{fig:hdrsun}
\end{center}
\vspace{-20pt}
\end{figure}
\textbf{Events to HDR image: }
%
In this paper, it is clearly shown that event stacks have rich information for HDR image reconstruction. In many cases, some parts of the scene are not visible in the APS image because of its low dynamic range. But many events really exist in those regions in the event camera as in the region under the table in Fig.~\ref{fig:impress} or the checkerboard pattern at the top left part of the stacked image in Fig.~\ref{fig:stream}. Although both examples are from dark illumination but normal cameras also fail in rather bright illumination. Figure~\ref{fig:hdrsun} shows the ability of the proposed method for HDR image generation in such cases. 

\textbf{Events to high frame rate video: }
The motion blur due to fast motion of a camera or the scene is one of the challenging problems, and this makes the vision methods unreliable. 
However, our method 
can actually generate very high frame rate (HFR) videos with much less motion blur under the fast motion as mentioned in Sec.~\ref{sec:video_stack}. To prove this ability, we conducted the tracking experiments using the reconstructed HFR video: with the event-based high frame rate video reconstruction framework, we can recover clear motion of a star-shape object attached on a fan with rotation speed of 13000 RPM, and the result shows that it is capable to generate the motion up to 1 million fps. \emph{The qualitative results will be shown in supplementary material.}  


\section{Conclusion} 
We demonstrated how our cGANs-based approach can benefit from the properties of event cameras to accurately reconstruct HDR non-blurred intensity images and high frame rate videos from pure events. 
We first proposed two initiative event stacking methods (SBT and SBE) for both image and video reconstruction from events using the network.
We then showed the advantages of using event cameras to generate high dynamic range images and high frame rate rate videos through experiments based on our datasets made of online available real-world sequences and simulator.
In order to show the robustness of our method, we compared our cGANs-based event-to-image framework with other existing reconstruction methods and showed that our method outperforms other methods on public available datasets. We also showed it is possible to generate high dynamic range images even in extreme illumination conditions and also non-blurred images under rapid motion. 

{\small
\bibliographystyle{ieee}
\bibliography{eventGANs}
}

\end{document}